\documentclass[runningheads]{llncs}
\usepackage{graphicx}

\usepackage{float}
\usepackage{amsmath}
\usepackage{amsfonts}
\usepackage[ruled,vlined,linesnumbered]{algorithm2e}
\usepackage{algpseudocode}
\usepackage{adjustbox}
\usepackage{caption}
\usepackage{subcaption}
\usepackage{listings}
\usepackage[usenames,dvipsnames]{xcolor}
\usepackage[toc,page]{appendix}

\usepackage[autoload=false,theme=grayscale]{jlcode}
\lstdefinelanguage{Julia}%
  {morekeywords={abstract,break,case,catch,const,continue,do,else,elseif,%
      end,export,false,for,function,immutable,import,importall,if,in,%
      macro,module,otherwise,quote,return,switch,true,try,type,typealias,%
      using,while},%
   sensitive=true,%
   alsoother={$},%
   morecomment=[l]\#,%
   morecomment=[n]{\#=}{=\#},%
   morestring=[s]{"}{"},%
   morestring=[m]{'}{'},%
}[keywords,comments,strings]%

\lstdefinestyle{Myjulia} {language=Julia,captionpos=b,tabsize=3,frame=lines,keywordstyle=\color[rgb]{0.6,0,0},morekeywords={or, then, s, End, c, cr, cn, cfr, cfn, xs, l, f, clear, bpull, proceed, cdr, cdn, o },numbers=left,showtabs=false,morecomment=[l]//, commentstyle=\color{gray}, xleftmargin=\parindent,
breaklines=true,stringstyle=\color[rgb]{0.627,0.126,0.941},
basicstyle=\footnotesize\ttfamily}

\begin{document}
\title{SeaPearl: A Constraint Programming Solver guided by Reinforcement Learning}
\titlerunning{SeaPearl: A CP Solver guided by RL}
%
\author{Félix Chalumeau\inst{1, \star} \and
Ilan Coulon\inst{1, \star}\ \and
Quentin Cappart\inst{2} \and Louis-Martin Rousseau\inst{2}}
\authorrunning{Chalumeau, Coulon et al.}
%
\institute{École Polytechnique, Institut Polytechnique de Paris, Palaiseau, France \\ \email{\{felix.chalumeau,ilan.coulon\}@polytechnique.edu}
\and
École Polytechnique de Montréal, Montreal, Canada
\\ \email{\{quentin.cappart,louis-martin.rousseau\}@polymtl.ca}}

\footnotetext[1]{*~The two authors contributed  equally to this paper.}
\maketitle              
\begin{abstract}

The design of efficient and generic algorithms for solving combinatorial optimization problems has been an active field of research for many years. Standard exact solving approaches are based on a clever and complete enumeration of the solution set. A critical and non-trivial design choice with such methods is the branching strategy, directing how the search is performed. 
The last decade has shown an increasing interest in the design of machine learning-based heuristics to solve combinatorial optimization problems. The goal is to leverage knowledge from historical data to solve similar new instances of a problem. Used alone, such heuristics are only able to provide approximate solutions efficiently, but cannot prove optimality nor bounds on their solution.
Recent works have shown that reinforcement learning can be successfully used for driving the search phase of constraint programming (CP) solvers. However, it has also been shown that this hybridization is challenging to build, as standard CP frameworks do not natively include machine learning mechanisms, leading to some sources of inefficiencies. This paper presents the proof of concept for \texttt{SeaPearl}, a new CP solver implemented in \textit{Julia}, that supports machine learning routines in order to learn branching decisions using reinforcement learning. Support for modeling the learning component is also provided.
We illustrate the modeling and solution performance of this new solver on two problems. Although not yet competitive with industrial solvers, \texttt{SeaPearl} aims to provide a flexible and open-source framework in order to facilitate future research in the hybridization of constraint programming and machine learning.

\end{abstract}
%
%
%



\section{Introduction}

The goal of combinatorial optimization is to find an optimal solution among a finite set of possibilities.
Such problems are frequently encountered in transportation, telecommunications, finance, healthcare, and many other fields \cite{anagnostopoulos2010portfolio,baptiste2012constraint,li2008modeling,schaus2009scalable,toth2002vehicle,tsolkas2012graph}.
Finding efficient methods to solve them has motivated research efforts for decades.  Many approaches have emerged in the recent years seeking to take advantage of learning methods to outperform standard solving approaches. Two types of approaches have been particularly successful while still showing drawbacks.
First,
machine learning approaches, such as deep reinforcement learning (DRL),
have shown their promise for designing good heuristics dedicated to solve combinatorial optimization problems \cite{bello2017neural,dai2018learning,deudoncournut2018,kool2018attention}.
The idea is to leverage knowledge from historical data related to a specific problem in order to solve rapidly future instances of the problem.
Although, a very fast computation time for solving the problem is guaranteed, 
such approaches only act as a heuristics and no mechanisms for improving a solution nor to obtain optimality proofs are proposed.
A second alternative is to embed a learning-component inside a search procedure. 
This has been proposed for mixed-integer programming \cite{gasse2019exact}, local search \cite{gambardella1995ant,zhou2016reinforcement}, SAT solvers \cite{selsam2019guiding,nejati2020machine}, and
constraint programming \cite{antuori2020leveraging,cappart2020combining}.
However, it has been shown that such a hybridization is challenging to build, 
as standard optimization frameworks do not natively include machine learning mechanisms,
leading to some sources of inefficiencies.
As an illustrative example, Cappart et al. \cite{cappart2020combining} 
used a deep reinforcement learning approach to learn the value-selection heuristic for driving the search of a CP solver.
To do so, they resorted to a \textit{Python} binding in order to call deep learning routines in the solver \texttt{Gecode} \cite{schulte2006gecode}, 
causing an important computational overhead.

Following this idea, we think that learning branching decisions in a constraint programming solver is an interesting research direction.
That being said, we believe that a framework that can be used for prototyping and evaluating new ideas is currently missing in this research field.
Based on this context, this paper presents \texttt{SeaPearl} (homonym of \texttt{CPuRL}, standing for \textit{constraint programming using reinforcement learning}), 
a flexible, easy-to-use, research-oriented, and open-source constraint programming solver able to natively use deep reinforcement
learning algorithms for learning value-selection heuristics. The philosophy behind this solver is to ease and speed-up the development process to any researcher desiring 
to design learning-based approaches to improve constraint programming solvers. 
Accompanying this paper, the code is available on Github\footnote{https://github.com/corail-research/SeaPearl.jl}, together with a tutorial showcasing 
the main functionalities of the solver and how specific design choices can be stated.
Experiments on two toy-problems, namely the \textit{graph coloring} and the 
\textit{travelling salesman problem with time windows}, are proposed in order 
to highlight the learning aspect of the solver. Note also that compared to \textit{impact-based search} \cite{refalo2004impact}, 
hybridization with ant colony optimization \cite{solnon2010ant}, or similar mechanisms where 
a learning component is used to improve the search 
of the solving process for a specific instance, the goal of our learning process is to leverage knowledge learned from other similar instances.
This paper is built upon the proof of concept proposed by Cappart et al. \cite{cappart2020combining}.
Our specific and original contributions are as follows: 
(1) we propose an architecture able to solve CP models, whereas \cite{cappart2020combining} was restricted to dynamic programming models,
(2) the learning phase is fully integrated inside the CP solver, and 
(3) the reinforcement learning environment is different as it allows CP backtracking inside an episode.
Finally, the solver is fully implemented in \textit{Julia} language, avoiding the overhead of \textit{Python} calls from a \texttt{C++} solver.

The next section introduces reinforcement learning and graph neural network, a
deep architecture used in the solver. The complete architecture of \texttt{SeaPearl} is then proposed, 
followed by an illustration of the modelling support for the learning component.
Finally, experiments and discussions about research directions that can be carried out with this solver are proposed.

\section{Technical Background}

This section gives details about the two main concepts that make \texttt{SeaPearl} different from other CP solvers.

\subsection{Reinforcement Learning}

\textit{Reinforcement Learning} (RL) \cite{sutton2018reinforcement} is a sub-field of machine learning dedicated to train agents to take actions in an environment in order to maximize an accumulated reward.
The goal is to let the agent interacts with the environment and discovers which sequences of actions lead to the highest reward. 
Formally, let $ \langle S, A, T, R \rangle$ be a tuple representing the environment, where $S$ is the set of states that can be encountered in the environment, $A$ is the set of actions that can be taken by the agent, $T : S \times A \mapsto S$ is the transition function  leading the agent from a state to another one given the action taken and, $R : S \times A \mapsto \mathbb{R}$ is the reward function associated with a particular transition. The behaviour of an agent is driven by its policy $\pi : S \mapsto A$, deciding which action to take when facing a specific state $S$. 
The goal of an agent is to compute a policy maximizing the accumulated sum of rewards during its lifetime, referred to as an \textit{episode}, and defined by a sequence of states
$s_t \in S$ with $t \in [1, T]$ and $s_T$ is the terminal state. Considering a discounting factor $\gamma$, the total return at step $t$ is denoted by $G_t = \sum_{k=t}^{T} \gamma^{k-t} R(s_k, a_k)$.

In deterministic environments, the value of taking an action $a$ from a state $s$ under a policy $\pi$ is defined by the action-value function $Q^{\pi}(s_t, a_t) = G_t$. Then, the problem consists in finding a policy that maximizes the final return: $\pi^* = \text{argmax}_{\pi} ~ Q^{\pi}(s, a)$, $\forall s, a \in S \times A$.
However, the number of possibilities has an exponential increase with the number of states and actions, which makes solving this problem exactly intractable.
Reinforcement learning approaches tackle this issue by letting the agent interact with the environment in order to learn information that can be leveraged to build a good policy.
Many RL algorithms have been developed for this purpose, the most recent and successful ones are based on deep learning \cite{goodfellow2016deep} and are referred to as \textit{deep reinforcement learning} \cite{arulkumaran2017deep}. The idea is to approximate either the policy $\pi$, or the action-value function $Q$ by a neural network
in order to scale up to larger state-action spaces. For instance, \textit{value-based methods}, such as DQN \cite{mnih2013playing}, have the following approximation: $\hat{Q}^{\pi}(\theta, s_t, a_t) \approx Q^{\pi}(s_t, a_t)$; whereas \textit{policy-based methods} approximates the policy: $\hat{\pi}(\theta,s) \approx \pi(s)$,  where $\theta$ are parameters of a trained neural network.

\subsection{Graph Neural Network}\label{part:mlgraph}

Learning on graph structures is a recent and active field of research in the machine learning community. It has plenty of applications such as molecular biology \cite{kearnes2016molecular}, social sciences \cite{monti2019fake}, and physics \cite{sanchez2020learning}. It has also been considered for solving combinatorial optimization problems \cite{cappart2021combinatorial}.
Formally, let $G = (V,E)$ be a graph with $V$ the set of vertices, $E$ the set of edges, $\textbf{f}_v\in \mathbb{R}^k$ a vector of $k$ features attached to a vertex $v\in V$, and similarly, $\textbf{h}_{v,u}\in \mathbb{R}^q$ a vector of $q$ features attached to an edge $(v,u)\in V$. 
Intuitively, the goal of \textit{graph neural networks} (GNN) is to learn a $p$-dimensional representation $\mu_v \in \mathbb{R}^p$ for each node $v \in V$ of $G$.
Similar to convolutional neural networks that aggregate information from neighboring pixels of an image, GNNs aggregate information from neighboring nodes using edges as conveyors. The features $f_v$ are aggregated iteratively with the neighboring nodes in the graph. After a predefined number of aggregation steps, the node embedding are produced and encompass both local and global characteristics of the graph.

Such aggregations can be performed in different ways. A simple one has been proposed by Dai et al.~\cite{dai2016discriminative} and used by Khalil et al.~\cite{khalil2017learning}. It works as follows.
Let $T$ be the number of aggregation steps, $\mu^t_v$ be the node embedding of $v$ obtained after $t$ steps and $\mathcal{N}(v)$ the set of neighboring nodes of $v\in V$ in $G$. The recursive computation of $\mu^t_v$ is shown in Eq. \eqref{eq:graph_embedding}, where vectors $\theta_1 \in \mathbb{R}^{p\times k}$, $\theta_2 \in \mathbb{R}^{p \times p}$, $\theta_3 \in \mathbb{R}^{p \times p}$, $\theta_4 \in \mathbb{R}^{p\times q}$ are vectors of parameters that are learned, and $\sigma$ a non-linear activation function such as \texttt{ReLU}. The final embedding $\mu_v^{T+1}$ obtained gives a representation for each node $v$ of the graph, that can consequently be used as input of regular neural networks for any prediction tasks.

\begin{equation}
\label{eq:graph_embedding}
\mu_v^{t+1} = \sigma \left(\theta_1 \textbf{f}_v + \theta_2 \sum_{u \in \mathcal{N}(v)}  \mu_u^t +
 \theta_3\sum_{u \in \mathcal{N}(v)}  \sigma \left(\theta_4 \textbf{h}_{v,u} \right) \right) \ \ \ \forall t \in \{1,\dots,T\}
\end{equation}

Many variants and improvements have been proposed to this framework. A noteworthy example is the \textit{graph attention network}  \cite{velivckovic2017graph}, that uses an attention mechanism \cite{bahdanau2014neural}, commonly used in recurrent neural networks. Detailed information about GNNs are proposed in the following surveys \cite{chami2021machine,Wu_2021,zhou2019graph,cappart2021combinatorial} and an intensive comparisons on the computational results of the different architectures have been proposed by 
Dwivedi et al. \cite{dwivedi2020benchmarking}.

\section{Embedding Learning in Constraint Programming}
\label{sec:embedding}

\begin{figure}[!ht]
\includegraphics[width=0.9\textwidth]{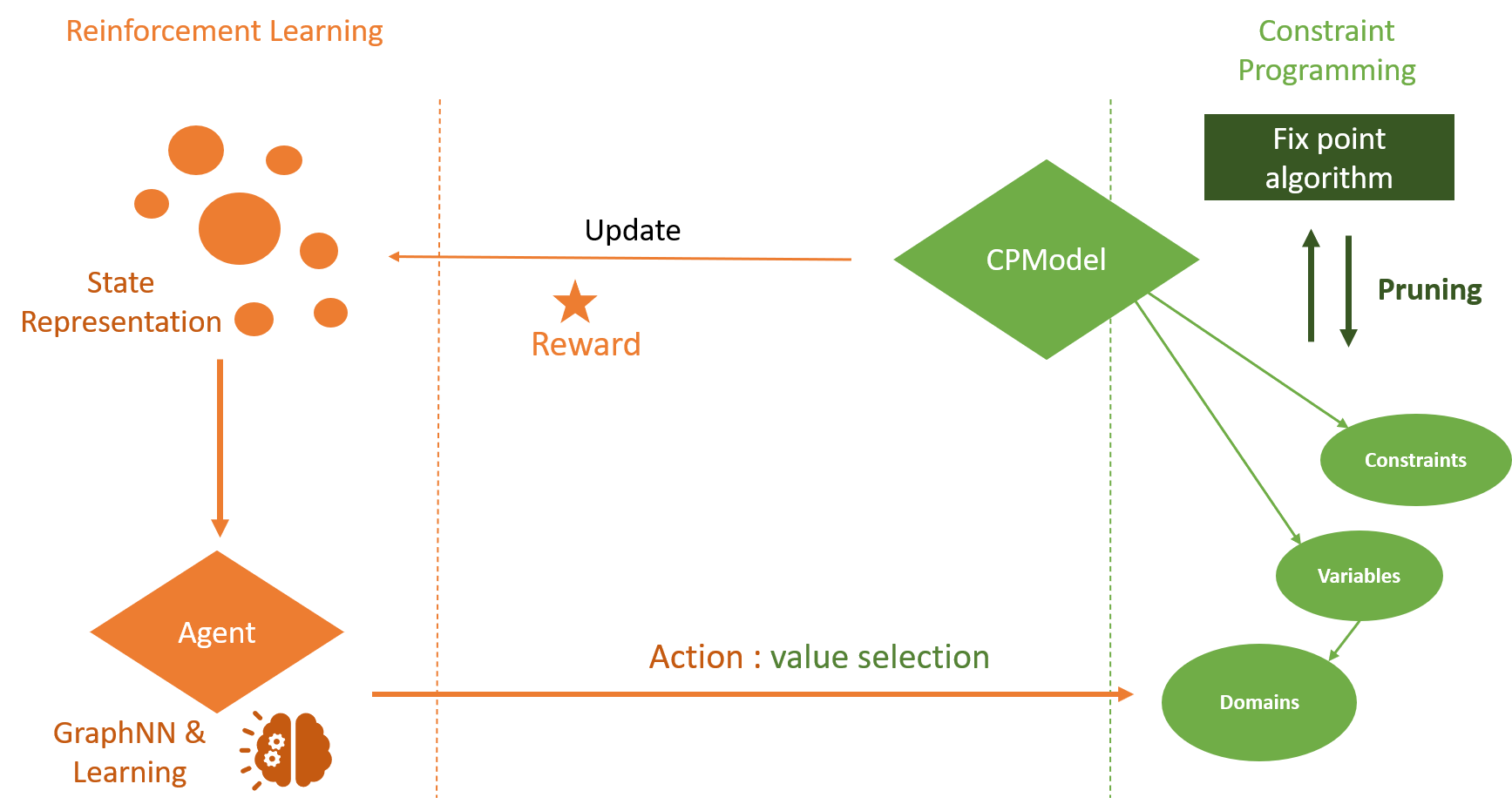}
\caption{Overview of \texttt{SeaPearl} architecture}
\label{fig:architecture}
\end{figure}

This section describes the architecture and the design choices behind \texttt{SeaPearl}.
A high-level overview is illustrated in Fig.~\ref{fig:architecture}.
Mainly inspired by \cite{cappart2020combining}, 
the architecture has three parts: 
a constraint programming solver, 
a reinforcement learning model,
and a common representation acting as a bridge between both modules.

\subsubsection*{Constraint Programming Solver}
A CP model is a tuple $\langle X, D, C, O \rangle$ where $X$ is the set of variables we are trying to assign a value to, 
$D(X)$ is the set of domains associated with each variable, 
$C$ the set of constraints that the variables must respect and 
$O$ an objective function. 
The goal of the solver is to assign a value for each variable $x \in X$ from $D(x)$ which satisfy all the constraints in $C$ and that optimize the objective function $O$.
The design of the solving process is heavily inspired by what has been done in modern \textit{trailing-based} solvers such as OscaR \cite{oscar}, or Choco \cite{jussien2008choco}.
It also takes inspiration from MiniCP \cite{minicp} in its philosophy.
The focus is on the extensibility and flexibility of the solver, especially for the learning component.
The goal is to make learning easy and fast to prototype inside the solver.
That being said, the solver is \textit{minimalist}. At the time of writing, only few constraints are implemented.

\subsubsection*{Reinforcement Learning Model}
The goal is to improve the CP solving process using knowledge from previously solved problems.
It is done by learning an appropriate value-selection heuristic and using it at each node of the tree search.
Following Bengio et al. \cite{bengio2020machine}, 
this kind of learning belongs to the third class (\textit{machine learning alongside optimization algorithms}) of ML approaches for solving combinatorial problems,
and raises many challenges.
To do so, a generic reinforcement learning environment genuinely representing the behaviour of the solving process for solving a CP model must be designed.
Let $\mathcal{Q}^p$ be a specific instance of a combinatorial problem $p$ we want to solve, $\mathcal{C}^p_i$ be the associated 
 CP model at the $i$-th explored node of the tree search, 
 and $\mathcal{S}^p_i$ be statistics of the solving process at the $i$-th node (number of bactracks, if the node has been already visited, etc.).
The environment $\langle S, A, T, R \rangle$ we designed is as follows.

\paragraph{State}
We define a state $s_i \in S$ as the triplet $(\mathcal{Q}^p, \mathcal{C}^p_i, \mathcal{S}^p_i)$.
By doing so, each state contains (1) information about the instance that is solved, (2) the current state of the CP model
and (3) the current state of the solving process.
In practice, each state is embedded into a $d$-dimensional vector of features, that serves as input for a neural network.
This can be done in different manners and two possible representations are proposed in the case studies.

\paragraph{Action} Each action corresponds to a value that can be assigned to a variable of the CP model.
An action $a \in A$ at a state $s_i \in S$ is available if and only if it is in the domain of the variable $x$ that has been selected
for branching on at step $i$ ($a \in D(X)$ for $ \mathcal{C}^p_i$).

\paragraph{Transition function}
The transition updates the current state according to the action that has been selected.
In our case, it updates the domains of the different variables.
It is important to highlight that the transition encompasses 
everything that is done inside the CP solver at each branching step, such as the constraint propagation during the fix-point computation, or the trailing in case of backtrack.
This is an important difference with \cite{cappart2020combining} where the transition only consists in the assignation of a value to a variable and is disconnected from the internal mechanisms of the CP solving process.

\paragraph{Reward function}
The reward is a key component of any reinforcement learning environment \cite{Dewey2014ReinforcementLA}.
In our case, it has a direct impact on how the tree search is explored.
Although it is commonly expressed as a function of the objective function $O$,
it is not clear how it can be shaped in order to drive the search to provide feasible solutions, the best one and to prove optimality, which often require different branching strategies.
For this reason, the solver allows the user to define its own reward.
That being said, a reward that gives a penalty of $-1$ at each step is integrated by default.
This simple reward encourages the agent to conclude an episode as soon as possible. 
The end of an episode means that the solver reached optimality. 
Hence, giving such a penalty drove the agent to reduce the number of visited nodes for proving optimality.
An alternative definition has been proposed in \cite{cappart2020combining}. The reward signal consists in two terms, having a different importance.
The first one is dedicated to find a feasible solution, whereas the second one drives the episode to find the best feasible solution.
The reward is designed in order to prioritize the first term. 
The motivation is to drive the search to find a feasible solution by penalizing the number of non-assigned variables before a failure, 
and then, driving it to optimize the objective function.

\paragraph{Learning agent}
Once the environment has been defined, any RL agent can be used to train the model \cite{mnih2013playing,schulman2017proximal}.
The goal is to build a neural network \texttt{NN} that outputs an appropriate value to branch on at each node of the tree search.
At the beginning of each new episode, an instance $\mathcal{Q}^p$ of the problem we want to solve is randomly taken from the training set and the learning is conducted on it.
The training algorithm returns a vector of weights ($\theta$) which is used for parametrizing the neural network. 
A recurrent issue in related works using learning approaches to solve combinatorial optimization problem is having access to enough data for training a model.
It is not often the case, and this makes the design of new approaches tedious.
To deal with this limitation, the solver integrates a support for generating synthetic instances, which are directly fed in the learning process.
Then, the training is done using randomly generated instances sampled from a similar distribution to those we want to solve.

\subsubsection*{State representation}

In order to ensure the genericity of the framework, 
the neural architecture must be able to take any triplet $(\mathcal{Q}^p, \mathcal{C}^p_i, \mathcal{S}^p_i)$ as input, which requires
to encode the RL state by a suitable representation.
Doing so for the statistics ($\mathcal{S}^p_i$) is trivial as they mainly consist of numerical values or categorical data.
The information related to the instances ($\mathcal{Q}^p$) is by definition problem-dependent, and several architectures are possible \cite{cappart2020combining}.
This information can also be omitted in our solver.
However, a representation able to handle any CP model ($\mathcal{C}^p_i$) is required.
In another context, Gasse et al. \cite{gasse2019exact} proposed a variable-constraint bipartite graph representation of mixed-integer linear programs 
in order to learn branching decisions inside a MIP solver using imitation learning. 
The representation they used is leveraged using a graph neural network. 
Following this idea, our solver adopted a similar architecture, referred to as a \textit{tripartite graph} but tailored for CP models.
A CP model $\langle X, D, C, O \rangle$ is represented by a simple and undirected graph $G(V_x,V_d,V_c,E)$ as follows.
Each variable $x \in X$ is associated to a vertex from $V_x$, each possible value (union of all the domains) to a vertex from $V_d$, 
and each constraint to a vertex from $V_c$. 
Edges only connect either nodes from $V_x$ to $V_c$ if the variable $x$ is involved in constraint $c$,  or nodes from $V_x$ to $V_d$  if $d$ is inside the current domain of $x$.
Finally, each vertex and each edge can be labelled with a vector of features, corresponding to additional information of the model (arity of a constraint, domain size of a variable, type of a global
constraint, etc.). 
The main asset of this representation is its genericity, as it can be used to represent any CP model.
It is important to note that designing the best state representation is still an open research question and many options are possible. 
Two representations have been tested in this paper. The first one is the generic representation based on the tripartite graph, whereas the second one leverages problem-dependent features, as in \cite{cappart2020combining}.

\begin{algorithm}[!ht]

$\triangleright$ \textbf{Pre:} $\mathcal{Q}^p$ is a specific instance of combinatorial problem $p$.

$\triangleright$ \textbf{Pre:} $\mathcal{C}^p_0$ is the state of the CP model $\langle X, D, C, O \rangle$ at the root node.

$\triangleright$ \textbf{Pre:} $\texttt{NN}$ is a neural architecture giving a value at a node of the tree.

$\triangleright$ \textbf{Pre:} $\textbf{w}$ is a trained weight vector parametrizing the neural network.

$\mathcal{C}^p_0 := \texttt{CPEncoding}(\mathcal{Q}_p)$

$\Psi := \texttt{CP-search}(\mathcal{C}^p_0)$

$i := 0$

	\While{$\Psi$ \textnormal{\textbf{is not completed}}}{
        
        $\texttt{fixPoint}(\mathcal{C}^p_i)$
        
        $\mathcal{S}^p_i := \texttt{getSearchStatistics}(\Psi)$
        
		$x := \texttt{selectVariable}(\mathcal{C}^p_i)$
		
		$v := \texttt{NN}(\textbf{w}, x, \mathcal{Q}^p, \mathcal{C}^p_i, \mathcal{S}^p_i)$
		
		$\mathcal{C}^p_{i+1} := \texttt{branch}(\Psi, x , v)$
		
		$ i := i +1 $
		
	}
\Return $\texttt{bestSolution}(\Psi)$

\caption{Solving process of \texttt{SeaPearl}}
\label{alg:seapearl}
\end{algorithm}

\subsubsection*{Solving Algorithm}

\begin{figure}[!ht]
\includegraphics[width=\textwidth]{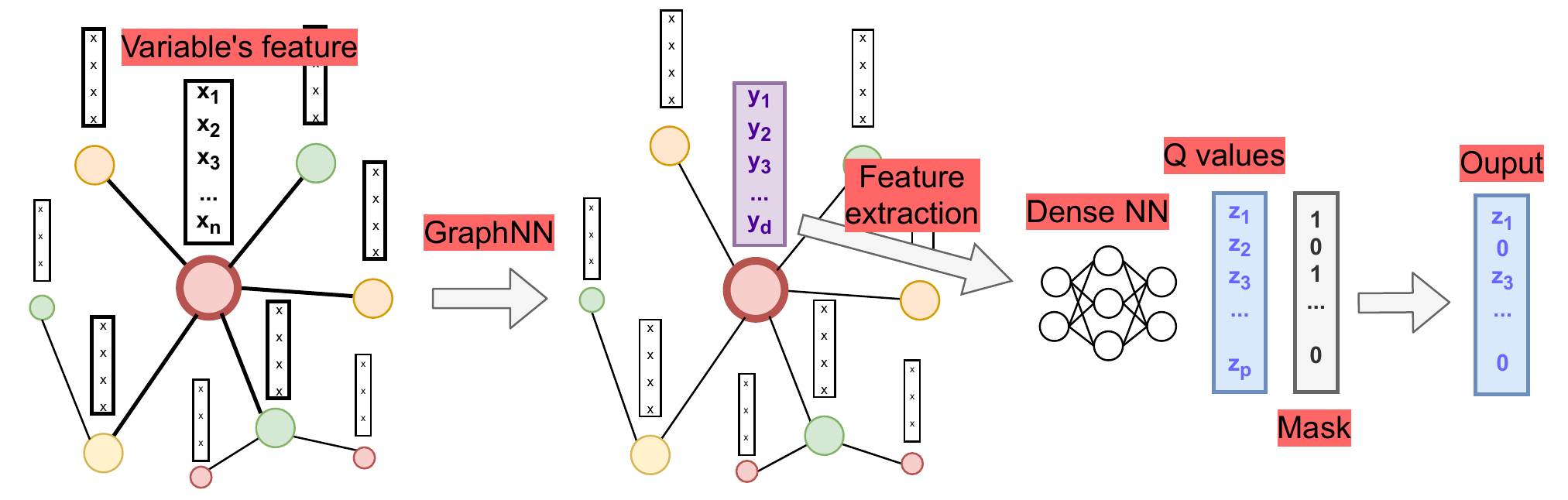}
\caption{Simplified representation of the neural network architecture}
\label{fig:messagepassing}
\end{figure}

The solving process of \texttt{SeaPearl} is illustrated in Algorithm \ref{alg:seapearl}.
It mostly works in the same manner as any modern CP solver.
The main difference is the consistent use of a learned-heuristic for the value selection.
While the search is not completed (lines 8-14), the fix-point algorithm is executed on the current 
node (line 9), and the features used as input of the neural network are extracted, both concerning the CP model (line 9) and the solving statistics (line 11).
Using such information, the trained model is called in order to obtain the value on which the current variable must be branched on (line 12).
Finally, the best solution found is returned (line 15). 
Note also that additional mechanisms, such as prediction caching \cite{cappart2020combining}, can be added to speed-up the search.
The architecture of the network considered (line 12) is proposed in Fig.~\ref{fig:messagepassing}. It works as follows:
(1) the GNN computes a latent $d$-dimensional vector representation of the features related to the current variable the tripartite graph, (2) the vector is used as input of a fully-connected neural network
in order to obtain a score for each possible value (resp. action), and (3) this score is passed through a mask in order to keep only the values that are inside the domain.

\section{Modeling, Learning and Solving with SeaPearl}
\label{ch:implementation}

The goal of \texttt{SeaPearl} is to make most of the previous building blocks transparent for the end-user.
This section illustrates how it is done with the graph coloring problem.
Let $G(V,E)$ be an undirected graph. A \textit{coloring} of $G$ is an assignment of labels to each node such that adjacent nodes have a different label. The \textit{graph coloring problem} consists in finding a coloring that uses the minimal number of labels.
The programming language that we selected to develop \texttt{SeaPearl} is \textit{Julia} \cite{Bezanson_2017},
which is (1) efficient during runtime, and (2) rich in both mathematical programming \cite{Dunning_2017,Distributions.jl-2019,Bromberger17,legat2020mathoptinterface} and machine learning libraries \cite{Flux.jl-2018,yuret2016knet}.
This resolves the issue encountered in \cite{cappart2020combining} where an inefficient \textit{Python} binding has been developed in order to call deep learning routines in \texttt{C++} implementation of the solver \texttt{Gecode} \cite{schulte2006gecode}. More examples are also available on the Github repository of the solver.
A regular CP model of the graph coloring problem is shown in Listing \ref{code:cp-model}.
The first step is to build a \textit{trailer}, instantiating the trailing mechanisms and a \textit{model}, instantiating the tuple $\langle X,D,C,O \rangle$.
Variables, constraints, and the objective are then added to this object. Once the model is built, the solving process can then be run. At that time, no learning is done.

\begin{jllisting}[language=julia, caption=CP model for the graph coloring problem, style=jlcodestyle, label=code:cp-model]
## Preamble ##
n_vertex, n_edge, edges = getInput(instance)
trailer = SeaPearl.Trailer()
model = SeaPearl.CPModel(trailer)

## Variable declaration ##
k = SeaPearl.IntVar(0, n_vertex, trailer)
x = SeaPearl.IntVar[]

for i in 1:n_vertex
    push!(x, SeaPearl.IntVar(1, n_vertex, trailer))
    SeaPearl.addVariable!(model, last(x))
end

## Constraints ##
for v1, v2 in input.edges
    push!(model.constraints, SeaPearl.NotEqual(x[v1], x[v2], trailer))
end

for v in x
    push!(model.constraints, SeaPearl.LessOrEqual(v, k, trailer))
end
    
## Objective (minimizing by default) ##
model.objective = k
SeaPearl.solve!(model)
\end{jllisting}

The next snapshot (Listing  \ref{code:nn}) shows how a reinforcement learning agent can be easily defined.
The first instruction corresponds to the definition of the deep architecture, consisting of 4 graph attention layers
(\texttt{GATConv}), and 5 fully-connected layers (\texttt{Dense}).
To do so, a binding with the library \texttt{Flux} \cite{Flux.jl-2018} has been developed.
The second instruction defines the deep reinforcement learning agent.
It requires as input (1) the neural architecture, (2) the optimizer desired (\texttt{Adam}), (3) the type of RL algorithm (\texttt{DQN})
and (4) hyper-parameters that depends on the specific algorithm selected (e.g., the discounting factor).

The last snapshot (Listing \ref{code:train}) shows how the training routines can be defined.
The value-selection heuristic is first defined as a heuristic that will be trained using the previously defined RL agent.
Then, a random generator, dedicated to construct new training instances, is instanciated.
For the graph coloring case, this generator is based on the construction proposed by \cite{Culberson95exploringthe}
and builds graphs of $n$ vertices with density $p$ that are $k$-colorable.
Finally, the training can be run. To do so, the user has to provide the value-selection to be trained, the instance generator,
the number of episodes, the search strategy, and the variable heuristic.
Once trained, the heuristic can be used to solve new instances.

\begin{jllisting}[language=julia, caption=Reinforcement Learning agent for the graph coloring problem, style=jlcodestyle, label=code:nn]
## Neural network architecture ##
neuralNetwork = SeaPearl.FlexGNN(
    graphChain = Flux.Chain(
        GATConv(nInput => 10, heads=2),
        GATConv(20 => 10, heads=3),
        GATConv(30 => 10, heads=3),
        GATConv(30 => 20, heads=2),
    ),
    nodeChain = Flux.Chain(
        Dense(20, 20),
        Dense(20, 20),
        Dense(20, 20),
        Dense(20, 20),
    ),
    outputLayer = Dense(20, nOutput))
                    
## Reinforcement learning agent ##
agent = RL.Agent(
    policy = RL.QBasedPolicy(
        learner = SeaPearl.CPDQNLearner(
            approximator = RL.Approximator(neuralNetwork, ADAM())
            loss_function = huber_loss,
            discounting_factor = 0.9999,
            batch_size = 32,
            ...
        ), 
        explorer = SeaPearl.CPEpsilonGreedyExplorer()))
\end{jllisting}

\begin{jllisting}[language=julia, caption=Training a value-selection heuristic, style=jlcodestyle, label=code:train]
## Defining the value selection heuristic as the RL agent ##
val_heuristic = SeaPearl.LearnedHeuristic(agent)

## Generating random instances ##
gc_generator = SeaPearl.GraphColoringGenerator()

## Training the model ##
SeaPearl.train!(
    valueSelectionArray = [val_heuristic], 
    generator = gc_generator,
    nb_episodes = 1000,
    strategy = SeaPearl.DFSearch,
    variableHeuristic = SeaPearl.MinDomain())
\end{jllisting}

We would like to highlight that these pieces of code illustrate only a small subset of the functionalities of the solver.
Many other components, such as the reward, or the state representation can be redefined by the end-user for prototyping new research ideas. 
This has been made possible thanks to the multiple dispatching functionality of \textit{Julia},
allowing the user to redefine types without requiring changes to the source code of \texttt{SeaPearl}.

\section{Experimental results}
\label{sec:exp}
The goal of the experiments is to evaluate the ability of \texttt{SeaPearl} to learn good heuristics for value-selection.
Comparisons against greedy heuristics on two NP-hard problems are proposed: \textit{graph coloring} and \textit{travelling salesman with time windows}. In order to ease the future research in this field and to ensure reproducibility, the implementation, the models and the results are
released in open-source with the solver.
Instances for training the models have been generated randomly with a custom generator. 
Training is done until convergence, limited to 13 hours on AWS' EC2 with 1 vCPU of Intel Xeon capped to 3.0GHz, and the memory consumption is capped to 32 GB.
The evaluation is done on other instances (still randomly generated in the same manner) on the same machine.

\subsection{Graph Coloring Problem}
\label{subsec:exp1}
\begin{figure}[!ht]
\centering
\begin{subfigure}[b]{0.49\textwidth}
     \centering
     \includegraphics[width=\textwidth]{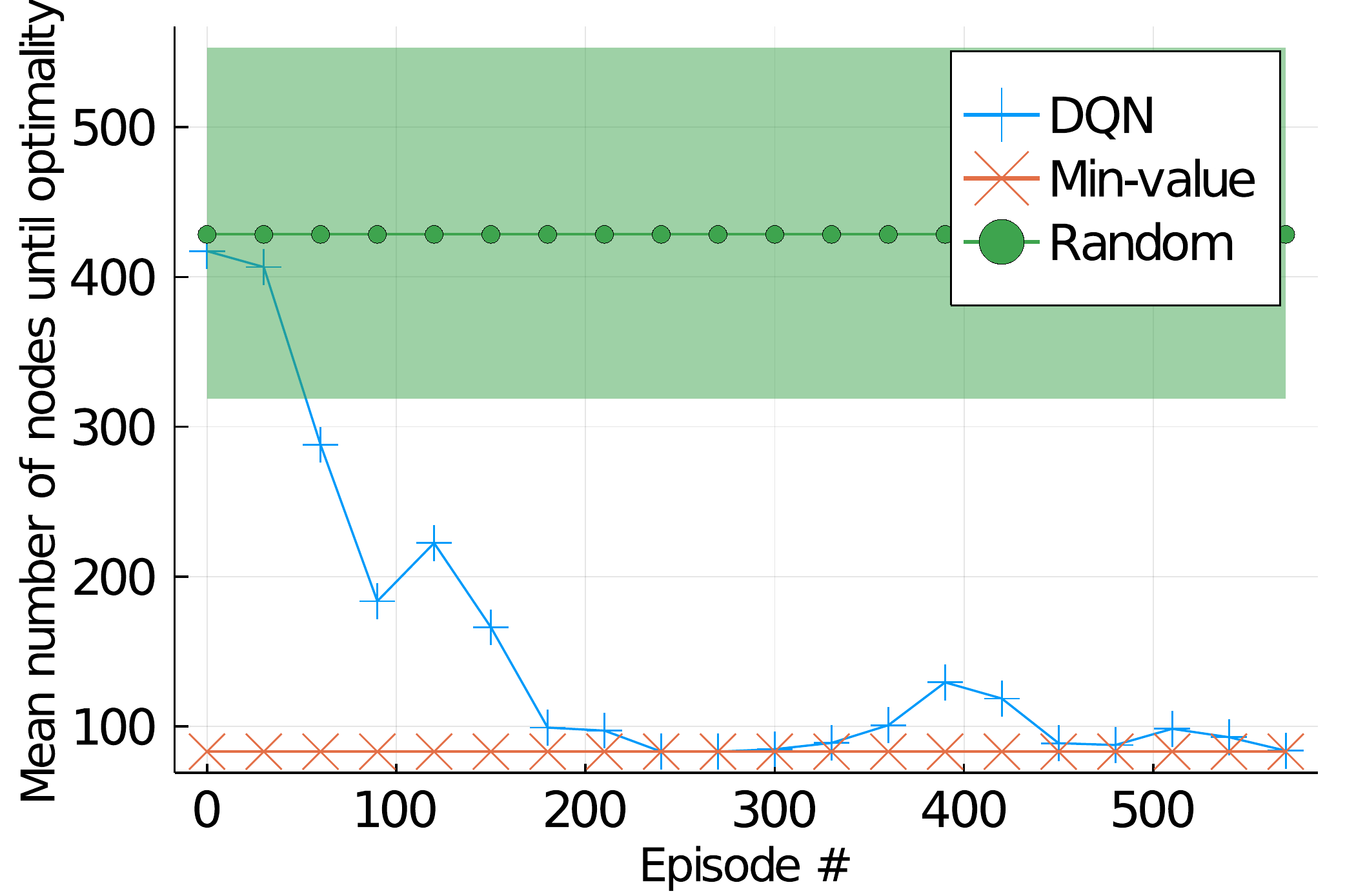}
     \caption{Instances with 20 nodes}
     \label{fig:dqngc20}
 \end{subfigure}
 \hfill
\begin{subfigure}[b]{0.49\textwidth}
     \centering
     \includegraphics[width=\textwidth]{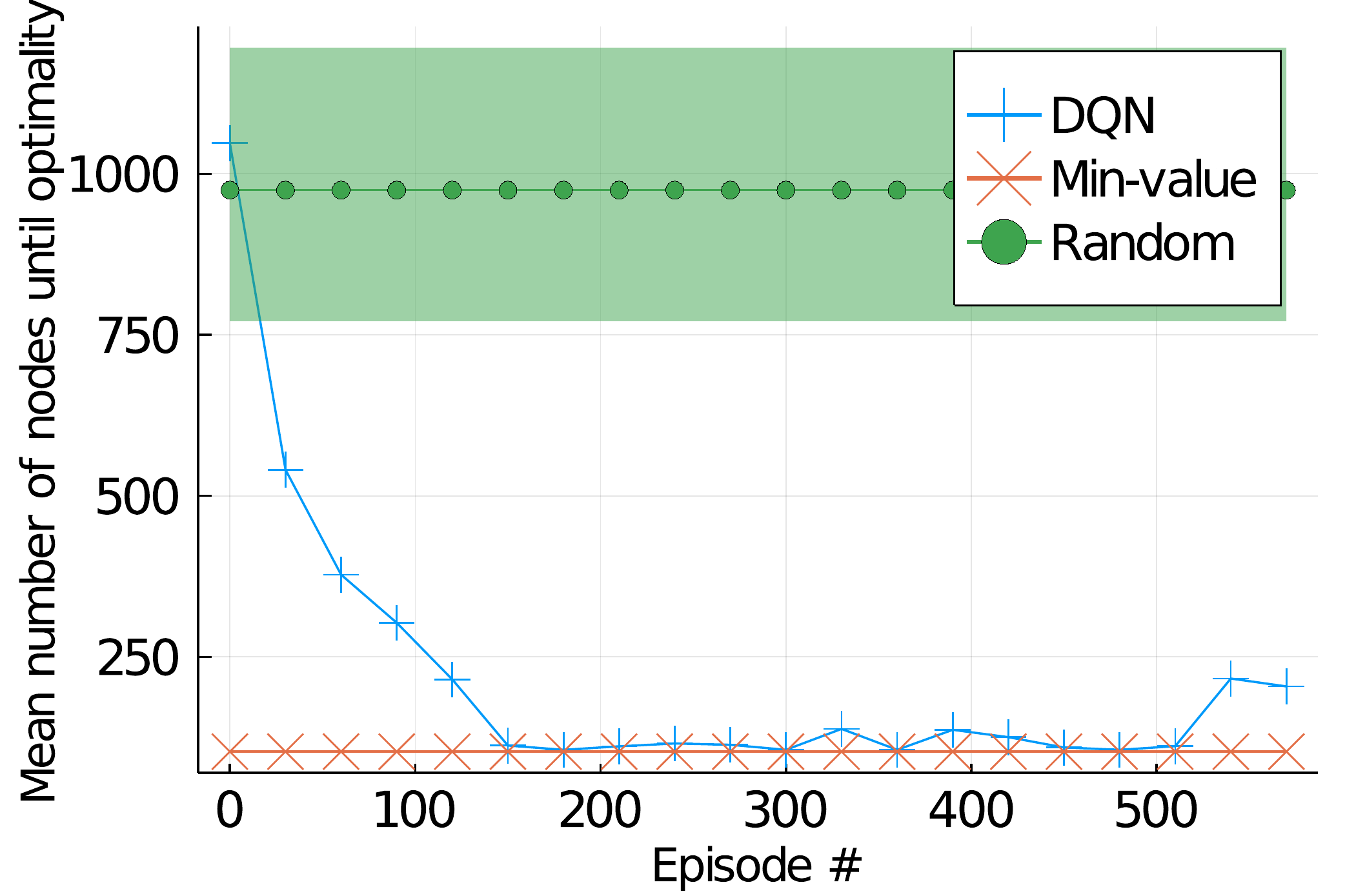}
     \caption{Instances with 30 nodes}
     \label{fig:dqngc30}
 \end{subfigure}
\caption{Training curve of the DQN agent for the graph coloring problem}
\label{fig:dqn_graphcoloring}
\end{figure}

\begin{figure}[!ht]
\centering
\begin{subfigure}[b]{0.49\textwidth}
     \centering
     \includegraphics[width=\textwidth]{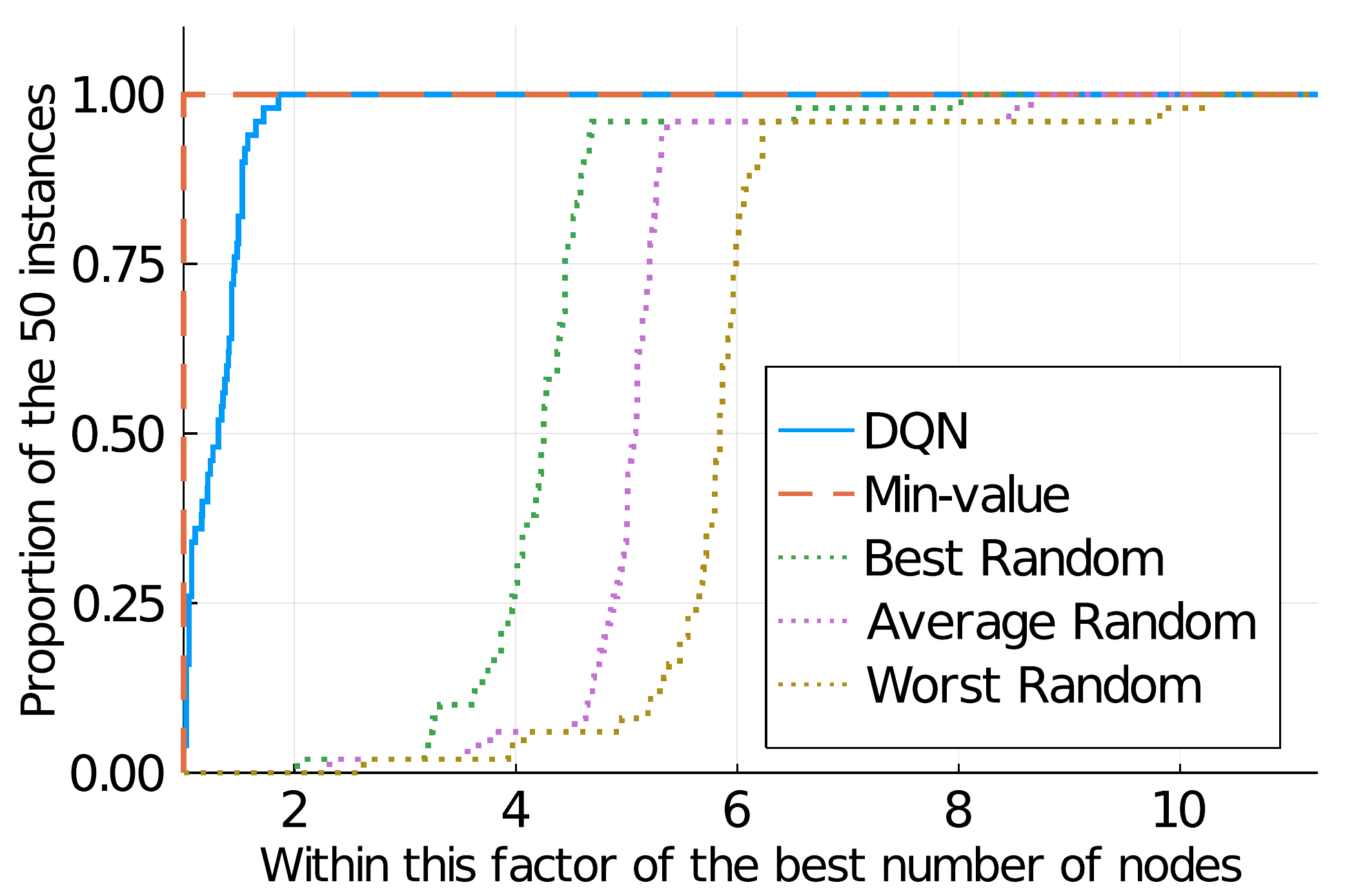}
     \caption{Instances with 20 nodes}
     \label{fig:20gc_profile_nodes}
 \end{subfigure}
 \hfill
\begin{subfigure}[b]{0.49\textwidth}
     \centering
     \includegraphics[width=\textwidth]{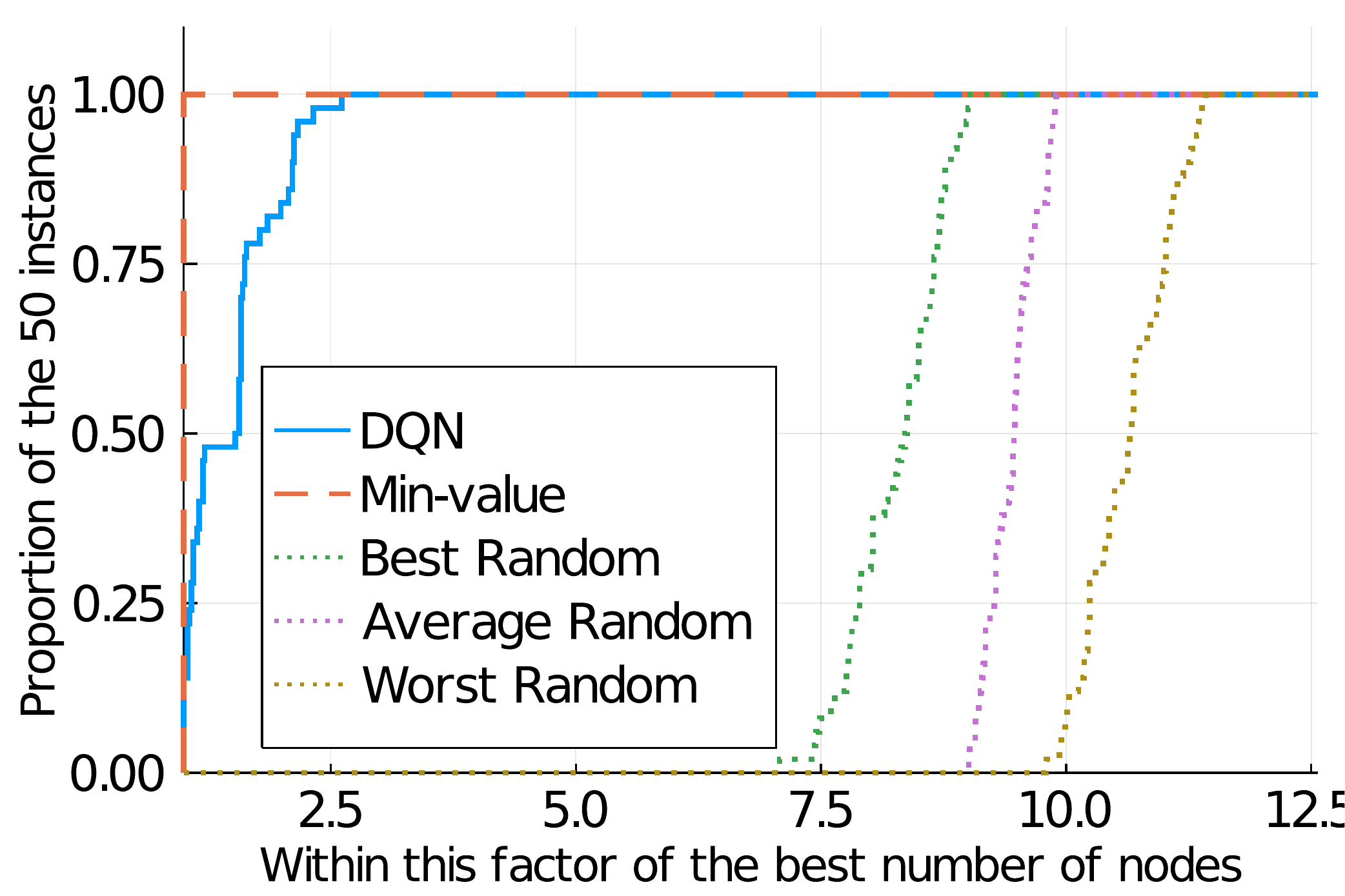}
     \caption{Instances with 30 nodes}
     \label{fig:30gc_profile_nodes}
 \end{subfigure}
 
\caption{Performance profiles (number of nodes) for the graph coloring problem}
\label{fig:gc_profile_nodes}
\end{figure}

 The experiments are based on a standard CP formulation of the graph coloring problem (Listing \ref{code:cp-model}), using the smallest domain as variable ordering. Instances are generated in a similar fashion as in \cite{Culberson95exploringthe}. 
They have a density of 0.5, and the optimal solutions have less than 5 colors. Comparisons are done with a heuristic that takes the smallest available label in the domain (min-value), and
a random value selection. For each instance, 200 random trials are performed and the average, best and worst results are reported. The training phase ran for 600 episodes (execution time of 13 hours) using DQN learning algorithm \cite{mnih2013playing}, a graph attention network has been used as deep architecture \cite{velivckovic2017graph} upon the tripartite graph detailed in Section \ref{sec:embedding}. 
A new instance is generated for each episode.
The first experiment records the average number of nodes that has been explored before proving the optimality of the instances at different steps of the training,
and using the default settings of \texttt{SeaPearl}.
Results are presented in Fig.~\ref{fig:dqn_graphcoloring} for graphs with 20 and 30 nodes. 
Every 30 episodes, an evaluation is performed on a validation set of 10 instances.
We can observe that the learned heuristic is able to reproduce the behaviour of the min-value heuristic, 
showing that the model is able to learn inside a CP solver.
Results of the final trained model on 50 new instances are illustrated in Fig.~\ref{fig:gc_profile_nodes} using performance profiles \cite{dolan2002benchmarking}.
The metric considered is still the number of nodes explored before proving optimality.
Results show that the heuristic performances can be roughly equaled.

\subsection{Travelling Salesman Problem with Time Windows}
\label{subsec:exp2}

Given a graph of $n$ cities, The \textit{travelling salesman problem with time windows} (TSPTW) consists of finding a minimum-cost circuit that connects a set of cities. Each city $i$ is defined by a position and a time window, defining the period when it can be visited. Each city must be visited once and the travel time between two cities $i$ and $j$ is defined by $d_{i,j}$.
It is possible to visit a city before the start of its time windows but the visitor must wait there until the start time.
However, it is not possible to visit a city after its time window. The goal is to minimize the sum of the travel distances.
This case study has been proposed previously as a proof of concept for the combination of constraint programming and 
reinforcement learning \cite{cappart2020combining}. We reused the same design choices they did. The CP model is based on a dynamic programming formulation,
the neural architecture is based on a graph attention network and the reward is shaped to drive the agent to find first a feasible solution, and then
to find the optimal one. It is also noteworthy to mention that the default tripartite graph of our solver is not used in this experiment.
A graph representing directly the current TSPTW instance is used instead, with the position and the time windows bounds as node features, and the distances
between each pair of nodes as edge features.

\begin{figure}[!ht]
\centering
\begin{subfigure}[b]{0.49\textwidth}
     \centering
     \includegraphics[width=\textwidth]{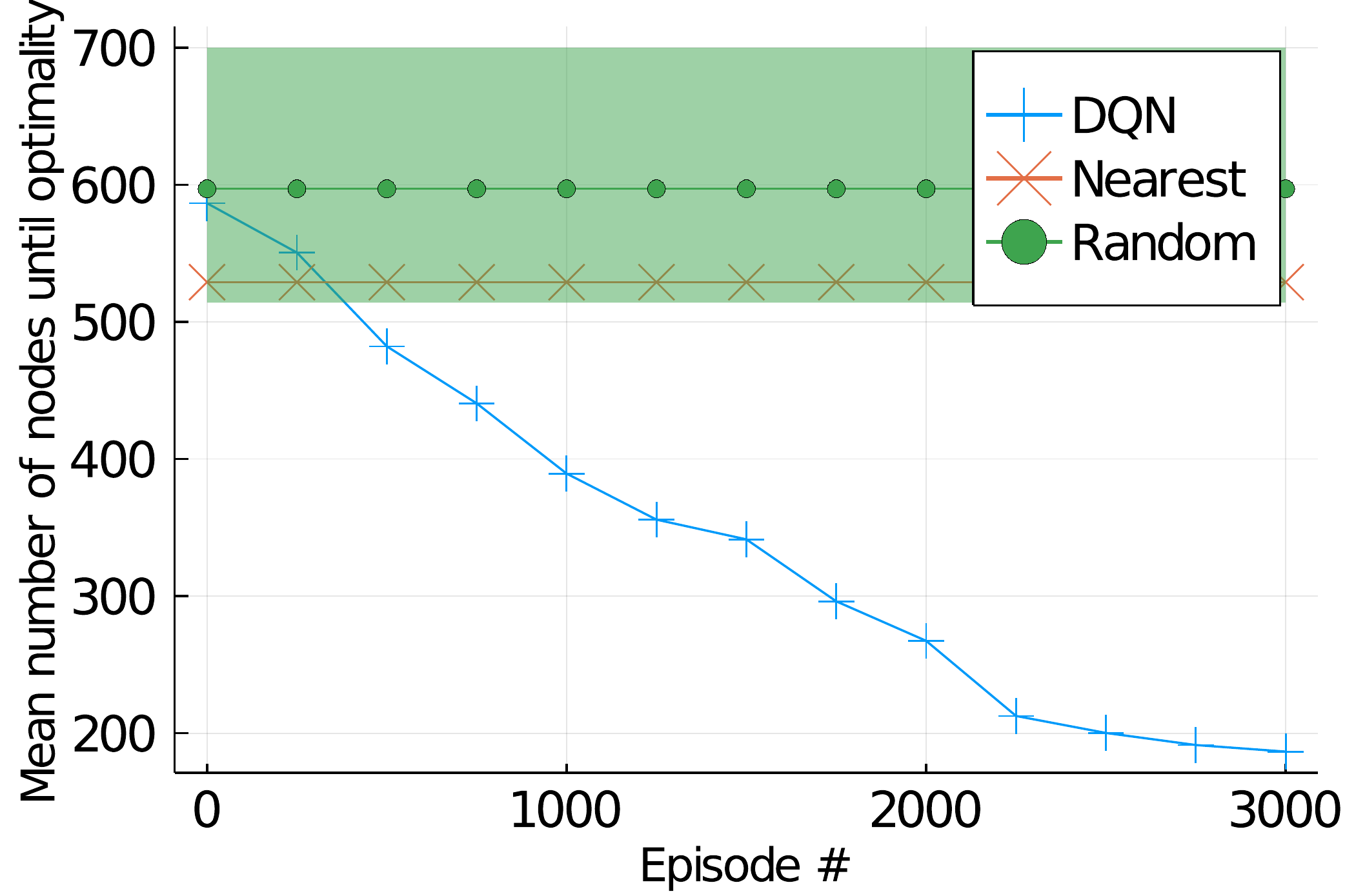}
     \caption{Instances with 20 cities}
     \label{fig:dqntsptw20}
 \end{subfigure}
 \hfill
\begin{subfigure}[b]{0.49\textwidth}
     \centering
     \includegraphics[width=\textwidth]{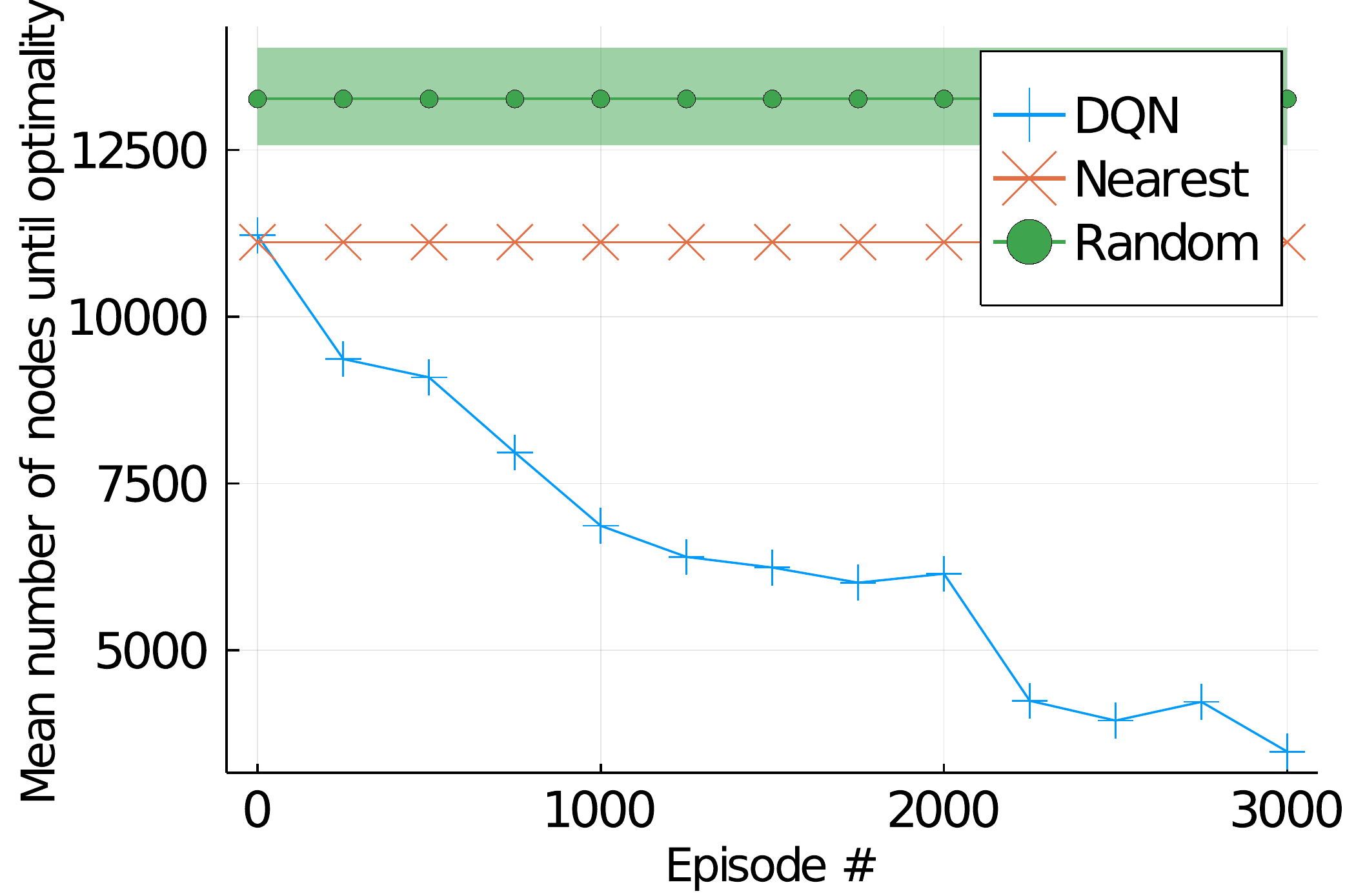}
     \caption{Instances with 50 cities}
     \label{fig:dqntsptw40}
 \end{subfigure}
\caption{Training curve of the DQN agent for the TSPTW}
\label{fig:dqntsptw}
\end{figure}

\begin{figure}[!ht]
\centering
\begin{subfigure}[b]{0.49\textwidth}
     \centering
     \includegraphics[width=\textwidth]{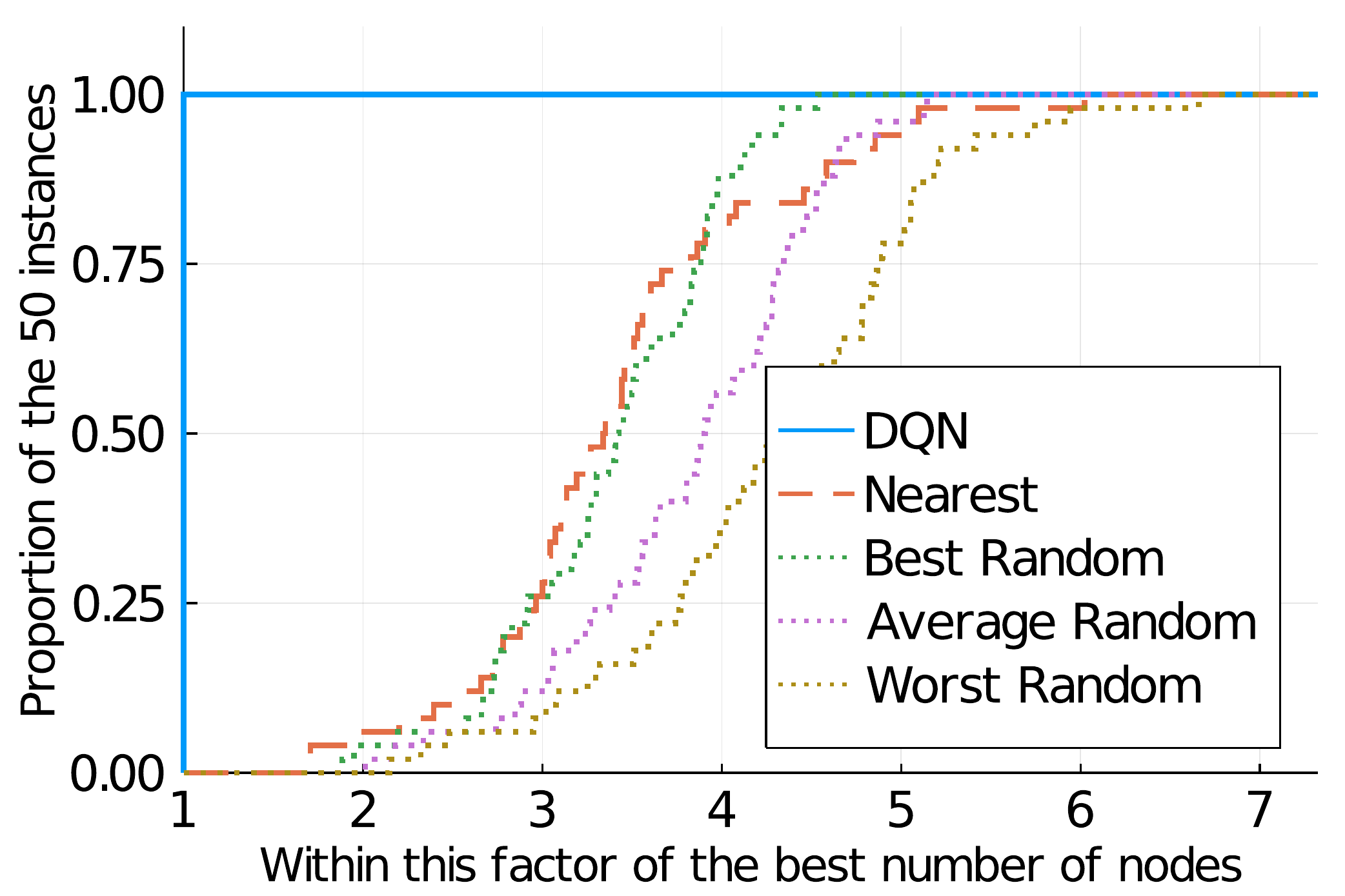}
     \caption{Instances with 20 cities (\# nodes)}
     \label{fig:20tsptw_profile-node}
 \end{subfigure}
 \hfill
\begin{subfigure}[b]{0.49\textwidth}
     \centering
     \includegraphics[width=\textwidth]{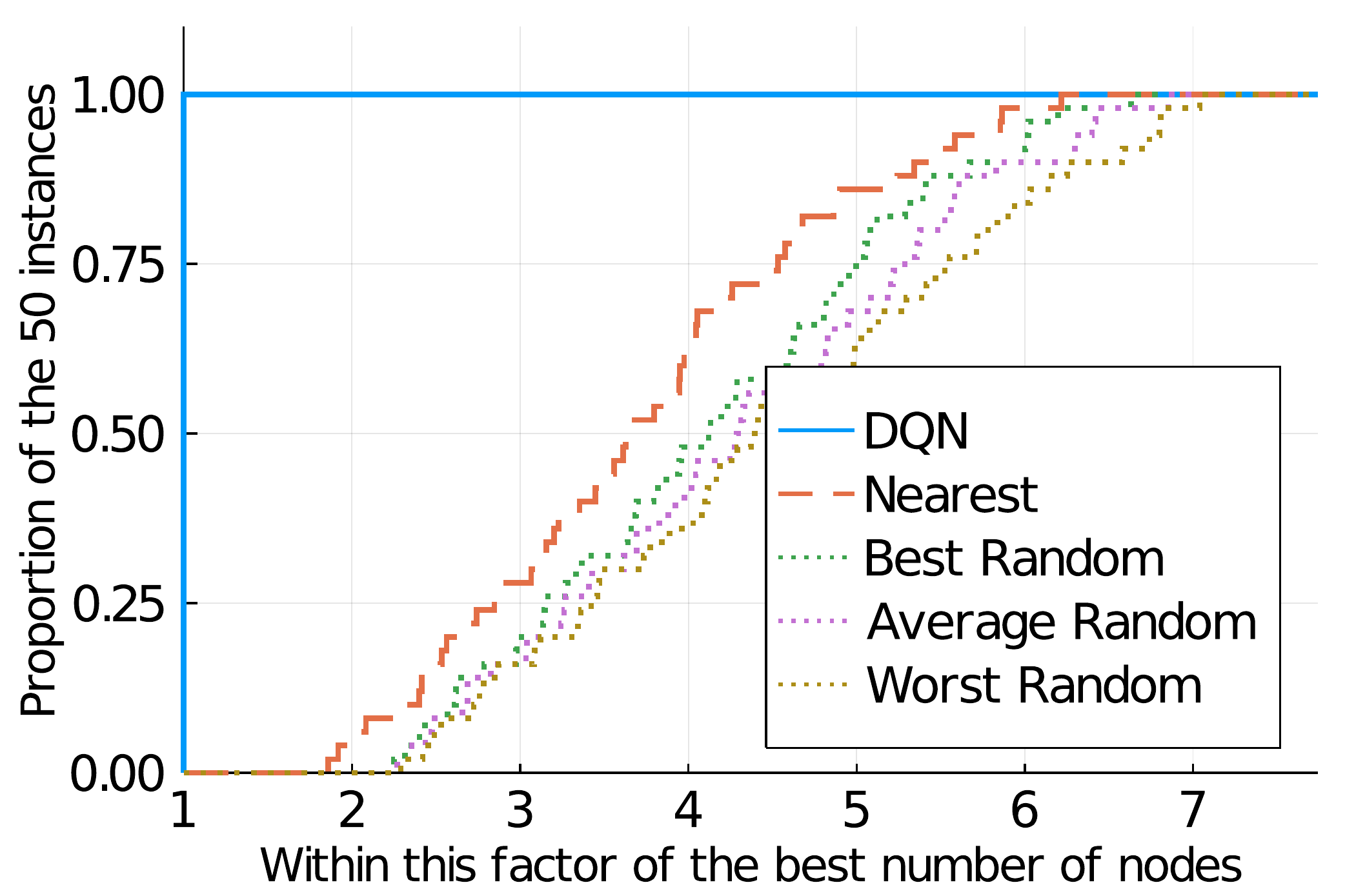}
     \caption{Instances with 50 cities (\# nodes)}
     \label{fig:50tsptw_profile-node}
 \end{subfigure}

\begin{subfigure}[b]{0.49\textwidth}
     \centering
     \includegraphics[width=\textwidth]{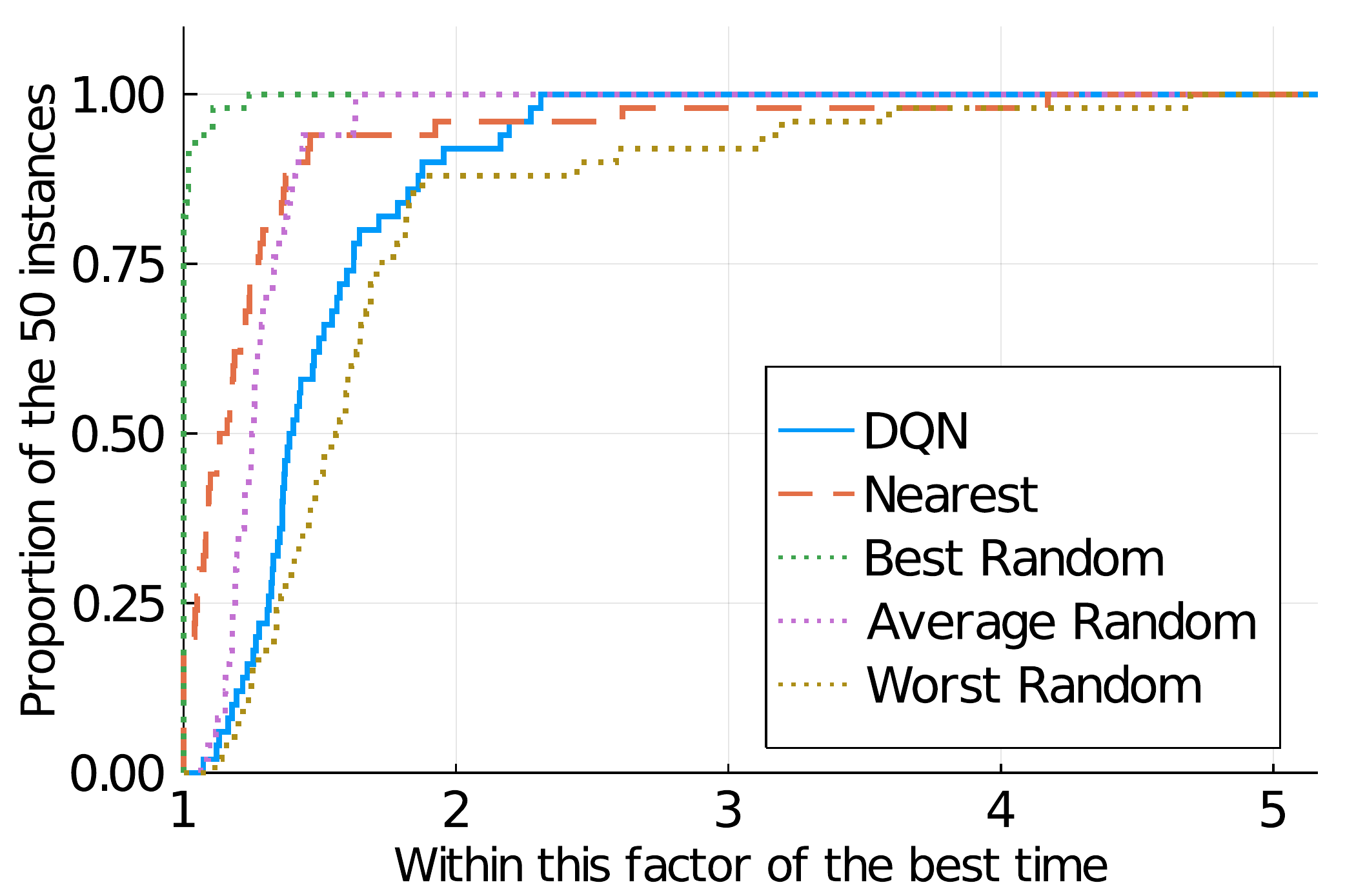}
     \caption{Instances with 20 cities (time)}
     \label{fig:20tsptw_profile-time}
 \end{subfigure}
 \hfill
\begin{subfigure}[b]{0.49\textwidth}
     \centering
     \includegraphics[width=\textwidth]{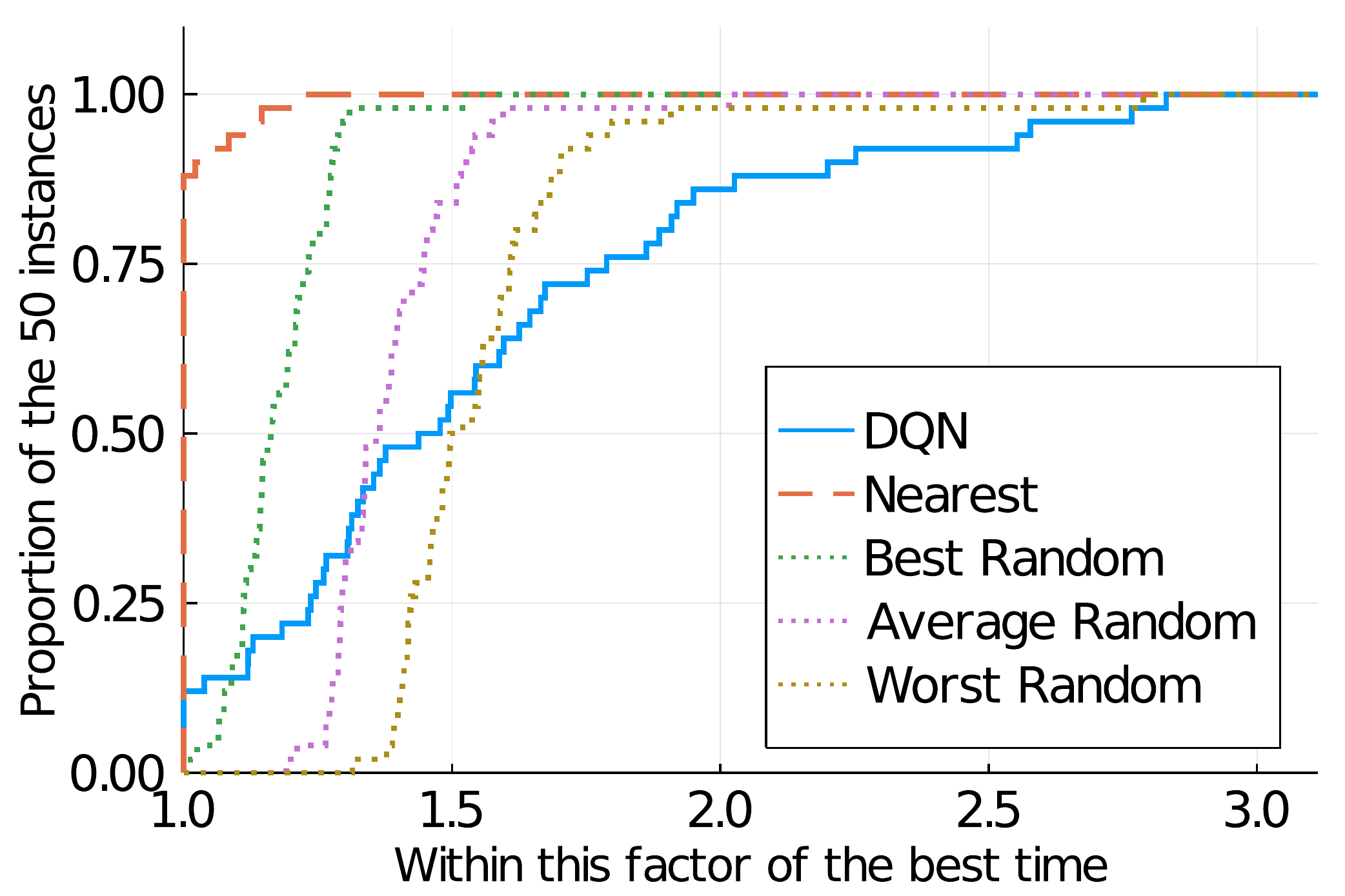}
     \caption{Instances with 50 cities (time)}
     \label{fig:50tsptw_profile-time}
 \end{subfigure}
\caption{Performance profiles for the TSPTW}
\label{fig:tsptw_profile}
\end{figure}

Instances are generated using the same generator as in \cite{cappart2020combining}. The training phase ran for 3000 episodes (execution time of 6 hours)
and a new instance is generated for each episode.
The variable ordering used is the one inferred by the dynamic programming model, and the value selection heuristic consists of taking the closest city to the current one.
The random value selection is also considered.
As with the previous experiment, we record the average number of nodes that have been explored before proving optimality.
Results are presented in Fig.~\ref{fig:dqntsptw} for instances with 20 and 50 cities. 
Once the model has been trained, we observe that the learned heuristic is able to outperform the heuristic baseline with a factor of 3 in terms of the number
of nodes visited. This result is corroborated by the performance profiles in Figs.~\ref{fig:20tsptw_profile-node}-\ref{fig:50tsptw_profile-node}, which shows the number of nodes explored before optimality
for the final trained model. 
Execution time required to solve the instances is illustrated in Figs.~\ref{fig:20tsptw_profile-time}-\ref{fig:50tsptw_profile-time}. 
We can observe that even if less nodes are explored, the greedy heuristic is still faster.
This is due to the time needed to traverse the neural network  at each node of the tree search.

\section{Perspectives and Future Works}

Leveraging machine learning approaches in order to speed-up optimization solver is a research topic that has an increasing interest \cite{cappart2020combining,antuori2020leveraging,gasse2019exact}.
In this paper, we propose a flexible and open-source research framework towards the hybridization of constraint programming and deep reinforcement learning. 
By doing so, we hope that our tool can facilitate the future research in this field.
For instance, four aspects can be directly addressed and experimented:  
(1) how to design the best representation of a CP state as input of a neural network, 
(2) how to select an an appropriate neural network architecture for efficiently learn value-selection heuristics,
(3) what kinds of reinforcement learning algorithm are the most suited for this task, and 
(4) how the reward should be designed to maximize the performances.
Besides, other research questions have emerged during the development of this framework.
This section describes five of them.

\subsubsection{Extending the Learning to Variable Selection}

As a first proof of concept, this work focused on how a value-selection heuristic can be learned inside a constraint programming solver.
Although crucial for the performances of the solver, especially for proving optimality  \cite{vilim_failure-directed_2015},
this has not been studied in this paper, and it has to be defined by the user. 
Integrating a learning component on it as well would be a promising direction.

\subsubsection{Finding Solutions and Proving Optimality Separately}\label{part:learningseparately}

As highlighted by Vilim et al. \cite{vilim_failure-directed_2015}, finding a good solution and exploring/pruning the search tree efficiently in order to prove optimality are two different tasks, that may require distinct heuristics. 
On the contrary, the reinforcement learning agent presented in this paper can hardly understand how and when both tasks should be prioritized.
This leads to another avenue of future work: 
having two specialized  agents, one dedicated to find good solutions, 
and the other one to prove optimality.

\subsubsection{Accelerating the Computation of the Learned Heuristics}

As for any solving tool, efficiency is a primary concern.
It is thus compulsory for the learned heuristic to be not only better than a man-engineered heuristic in terms of number of nodes visited 
but also in terms of execution time. 
As highlighted in the experiments,
calling a neural neural network is time consuming compared to calling a simple greedy heuristic.
Interestingly, this aspect has not been so considered in most deep learning works,
as the trained model are only called few times, rendering the inference time negligible in practice.
In our case, as the model has to be called at each node of the tree search (possibly more than 1 million times), the inference time becomes a critical concern.
This opens another research direction: finding an appropriate trade-off between a large model having accurate prediction and
a small model proposing worse prediction, but much faster. 
This has been addressed by Gupta et al. \cite{gupta2020hybrid} for standard MIP solvers.
Another direction is to consider the \textit{network pruning} literature, dedicated to reduce heavy inference costs of deep models in low-resource settings \cite{lin2017runtime,liu2018rethinking}.

\subsubsection{Reducing the Action Space}

A recurrent difficulty is to deal with problems having large domains.
On a reinforcement learning perspective, it consists of having a large action space, which makes the learning more difficult, 
and reduces the generalization to large instances.
A possible direction could be to reduce the size of the action space using a dichotomy selection.
Assuming a domain of $n$ values and a number of actions capped at $k$, 
the current domain is divided into $k$ intervals, and the selection of an action consists in taking a specific interval.
This can be done until a final value has been found. 
Another option is to use another architectures that are less sensitive to large action spaces, such as \textit{pointer networks} \cite{vinyals2015pointer}, which are commonly used in natural language processing but which have also been considered in combinatorial optimization \cite{deudoncournut2018}.

\subsubsection{Tackling real instances}
The data used to train the models are randomly generated from a specified distribution.
Although this procedure is common in much of published research in the field \cite{kool2018attention,gasse2019exact}, it cannot be used for solving real-world instances.
One additional difficulty to consider in real-world instances is having access to enough data to be able to accurately learn the distribution.
One way to do that is to modify slightly the available instances by introducing small perturbations on them. This is referred to as \textit{data augmentation},
but may be insufficient as it can fail to represent the distribution of the future instances.

\section{Conclusion}

Combining machine learning approaches with a search procedure in order to solve combinatorial optimization problems is a hot topic in the research community,
but it is still a challenge as many issues must be tackled.
We believe that the combination of constraint programming and reinforcement learning is a promising direction for that.
However, developing such hybrid approaches requires a tedious and long development process \cite{cappart2020combining}.
Based on this context, 
this paper proposes a flexible, easy-to-use and open-source research framework towards the hybridization of constraint programming and deep reinforcement learning.
The integration is done on the search procedure, where the learning component is dedicated to obtain a good value-selection heuristic.
Experimental results show that a learning is observed, and highlight challenges related to execution time.
Many open challenges should be addressed for an efficient use of machine learning methods inside a solving process. 
We position this contribution not only as a new CP solver,
but also as an open-source tool dedicated to help the community in the development of new hybrid approaches for tackling such challenges.

\bibliographystyle{splncs04}
\bibliography{ref}


\end{document}